\documentclass[sigconf]{acmart}

\usepackage{graphicx}
\usepackage{booktabs}
\usepackage{tabularx}
\usepackage{algorithm}
\usepackage{algpseudocode}
\usepackage{placeins}
\usepackage{dblfloatfix}

\setcopyright{none}
\renewcommand\footnotetextcopyrightpermission[1]{}
\acmConference[Preprint]{Preprint}{October 2026}{arXiv}
\graphicspath{{figures/}}
\AtBeginDocument{%
  \setlength{\abovedisplayskip}{3pt plus 1pt minus 1pt}%
  \setlength{\belowdisplayskip}{3pt plus 1pt minus 1pt}%
  \setlength{\abovedisplayshortskip}{2pt plus 1pt minus 1pt}%
  \setlength{\belowdisplayshortskip}{2pt plus 1pt minus 1pt}%
}
\algrenewcommand{\algorithmicrequire}{\textbf{Input:}}
\algrenewcommand{\algorithmicensure}{\textbf{Output:}}

\title{Zero-Shot Multi-Class Wafer Map Classification with Rule-Ordered Prompts for Large Multimodal Models}
\author{Tianjian Cao}
\affiliation{\institution{Fudan University}\city{Shanghai}\country{China}}
\email{tjcao24@m.fudan.edu.cn}
\author{Jun Tao}
\affiliation{\institution{Fudan University}\city{Shanghai}\country{China}}
\email{taojun@fudan.edu.cn}

\begin{document}

\begin{abstract}
Wafer-map defect classification often requires labeled data and retraining when production conditions change. We present a training-free six-class framework that combines connected-component highlighting with morphology-, location-, and exclusion-aware prompts for a frozen large multimodal model. On Mixed-WM38 (6,866 maps), multi-region prompting achieves 92.9\% macro and 93.8\% overall accuracy. On WM811K (9,000 maps), it achieves 79.1\% and 81.3\%, respectively. Eight alternative rule orders achieve 90.1--91.9\% overall accuracy, compared with 93.8\% for the selected order.
\end{abstract}

\keywords{wafer map classification, zero-shot learning, large multimodal model, prompt design, defect pattern recognition, semiconductor manufacturing}

\maketitle

\section{Introduction}

Wafer maps use a two-dimensional grid to record the spatial distribution of tested dies on a semiconductor wafer, and they are one of the most common quality representations in the manufacturing test stage. Each die is typically encoded as pass or fail and arranged according to its physical location on the wafer \cite{yuan2011,wu2015}. Recognizing spatial defect patterns in wafer maps is important for yield analysis, process diagnosis, and production control. From a process perspective, different manufacturing deviations can create different spatial failure patterns on the wafer plane. For example, chemical-mechanical planarization (CMP) non-uniformity may produce center-region failure aggregation. Layer-to-layer misalignment in the storage-node process may generate edge patterns along the wafer boundary, and localized non-uniformity or uneven cleaning can lead to local defect clusters \cite{kyeong2018}. Therefore, recognizing these specific patterns is practically valuable for translating wafer test results into actionable process feedback.

Existing automatic wafer map classification methods can be broadly grouped into unsupervised and supervised methods. Unsupervised methods extract spatial structures from unlabeled wafer maps using principal component analysis \cite{hsieh2010}, clustering, and spatial filtering \cite{ezzat2021}. They are useful for discovering latent failure modes, but are often sensitive to preprocessing and handcrafted feature design. Supervised methods learn class decision boundaries from labeled wafer maps. Earlier studies apply linear discriminant analysis \cite{yu2015} and support vector machine \cite{baly2012} to map handcrafted or statistical features to defect classes. CNN-based methods and deep models for mixed-type patterns later improve spatial feature learning \cite{nakazawa2018,kyeong2018}. Recent studies explore few-shot/self-supervised learning \cite{geng2021}, and state-space models for low-data settings or complex pattern modeling \cite{nie2025}. Although these methods can achieve strong performance on a given dataset, they assume that feature representations and decision boundaries are obtained through training samples, task adaptation, or model learning.

In real manufacturing environments, supervised methods face three major limitations. First, high-quality wafer maps require tape-out, testing, and expert annotation, while new production lines or process nodes may have few historical samples \cite{chen2007}. Second, changes in production line, process node, wafer size, or product design can shift defect distributions and require retraining or calibration \cite{zhao2022}. Third, human labels may contain noise and inconsistency, degrading model reliability \cite{frenay2014}.

Large multimodal models provide an alternative route to target-domain training. CLIP demonstrates zero-shot image classification through natural-language supervision \cite{radford2021}, while instruction-following LMMs such as GPT-4o support direct image-text reasoning \cite{hurst2024}. Unlike prompt learning \cite{zhou2022} and visual prompt tuning \cite{jia2022}, which optimize task-specific parameters, visual prompting uses visible marks or annotations to guide a frozen model toward relevant image regions \cite{yang2023fine,yang2023set}. Recent wafer-map research further shows that connected-component highlighting can support training-free binary normal/abnormal recognition with a frozen LMM \cite{cao2026}.

Building on the connected-component visual interface of \cite{cao2026}, which is developed for binary normal/abnormal recognition, this work studies fine-grained six-class classification. The main challenge is no longer only to locate salient defect regions, but to resolve overlapping class boundaries using morphology-, location-, and exclusion-aware textual rules. We formulate the task as a training-free semantic decision process in which the wafer map provides visual input and a fixed rule-presentation sequence specifies defect morphology, location, and class boundaries. This setting complements supervised classification by targeting the stage before representative target-domain data become available, and it supports inspection and revision of the rules without parameter retraining.

\begin{figure}[t]
  \centering
  \includegraphics[width=0.70\columnwidth]{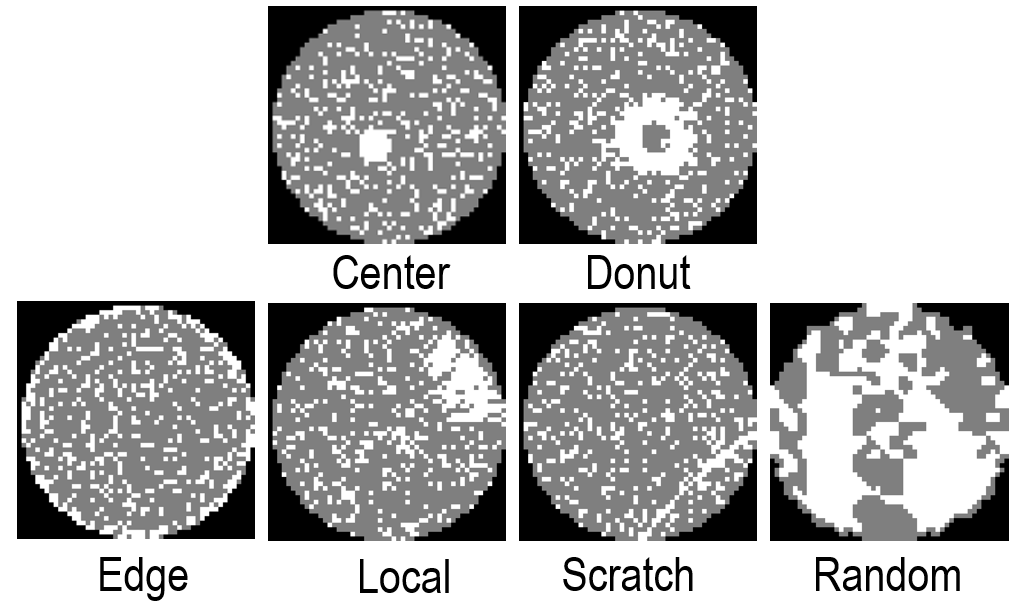}
  \caption{Representative raw wafer maps of the six target defect patterns from Mixed-WM38 \cite{wang2020}. White pixels denote failed dies, gray pixels denote non-defect wafer areas, and black regions denote the background.}
  \Description{Representative raw wafer maps for Center, Donut, Edge, Local, Random, and Scratch defect patterns.}
  \label{fig:patterns}
\end{figure}

We consider zero-shot classification of six wafer defect patterns: Center, Donut, Edge, Local, Random, and Scratch, as illustrated in Figure~\ref{fig:patterns}. Their recognition requires interpretation of both the morphology of dominant defect regions and their spatial organization across the wafer.

The proposed framework couples a connected-component-based visual representation with an ordered set of textual decision rules. The semantic rules map observed regions and distributions to defect classes, and the presentation order encodes an intended priority when multiple class descriptions are partially compatible with the same wafer map. Together, these components provide an inspectable, training-free decision process. The main contributions are:
\begin{itemize}
  \setlength{\topsep}{2pt}
  \setlength{\itemsep}{0pt}
  \setlength{\parsep}{0pt}
  \item We formulate six-class wafer-map recognition as a training-free image-text semantic decision problem and evaluate a frozen LMM on two industrial datasets.
  \item We introduce boundary-aware semantic rules that combine positive morphology, spatial anchoring, and exclusion conditions for neighboring classes.
  \item We define a fixed rule-presentation order that encodes an intended priority for overlapping evidence and evaluate controlled changes in regional marking, semantic detail, and rule order.
\end{itemize}

\section{Proposed Method}

\subsection{Overview}
\label{sec:overview}

\begin{figure}[t]
  \centering
  \includegraphics[width=0.95\columnwidth]{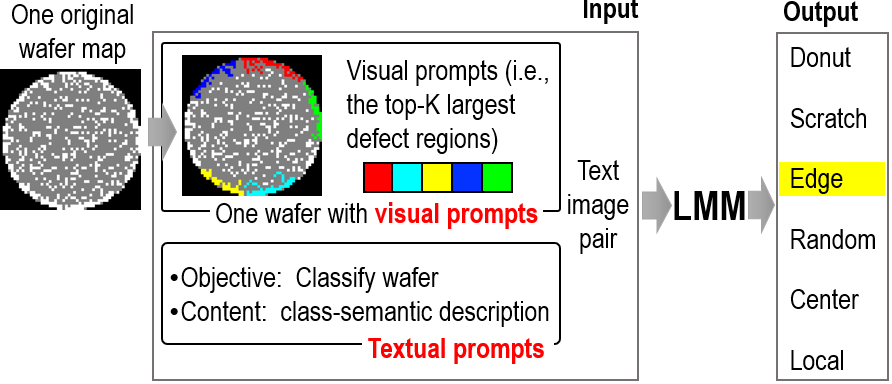}
  \caption{Overview of the training-free image-text classification framework. The visual input is paired with a textual prompt, and a frozen LMM processes the image-text pair once and returns exactly one label.}
  \Description{A workflow combining a visually prompted wafer map and ordered textual class rules as input to a frozen LMM, which returns one of six defect labels.}
  \label{fig:overview}
\end{figure}

We formulate multi-class wafer map classification as a training-free semantic decision problem. We consider six wafer defect patterns defined by their spatial morphology and location: Center, Donut, Edge, Local, Random, and Scratch, as illustrated in Fig.~\ref{fig:patterns}. Center denotes a defect cluster near the wafer center; Donut denotes a ring-like or hollow structure around the center; Edge denotes defect aggregation along the boundary; Local denotes compact local clusters; Scratch denotes an elongated defect with long-axis continuity; and Random denotes large block-like failures distributed across much of the wafer interior, leaving little continuous gray area. Because several classes share spatial cues---for example, Center and Donut are both center-related, while Edge and Local may both occur near the wafer boundary---we formulate recognition as a boundary-aware semantic decision problem. Let $X$ denote a raw wafer map and let
\begin{equation}
\mathcal{Y}=\{\mathrm{Center},\mathrm{Donut},\mathrm{Edge},\mathrm{Local},\mathrm{Random},\mathrm{Scratch}\}.
\label{eq:label-set}
\end{equation}
The visual prompting interface transforms $X$ into a visual representation $Z=V(X)$, while the textual prompt encodes a set of class-semantic rules $\mathcal{R}=\{r_c\mid c\in\mathcal{Y}\}$ and their presentation order $\pi$. A frozen LMM then predicts
\begin{equation}
\hat{y}=F_\theta\!\left(Z,P(Z;\mathcal{R},\pi)\right),
\qquad \hat{y}\in\mathcal{Y},
\label{eq:prediction}
\end{equation}
where $\theta$ remains fixed throughout inference. This formulation separates the visual evidence $Z$ from the semantic decision specification $P$, while allowing the textual legend to remain consistent with the visible form of $Z$. Figure~\ref{fig:overview} summarizes the resulting image-text classification framework.

Failed dies naturally form connected spatial structures on the wafer plane. Following \cite{cao2026}, $V$ uses these connected components as region-level visual references, allowing textual concepts such as ``dominant region,'' ``rim-following arc,'' and ``global distribution'' to be linked to visible regions in the image. Black pixels denote the background, gray pixels denote non-defect wafer areas, and failed dies constitute the defect mask. The mask is partitioned using 8-connectivity \cite{wu2009}, and the top-$K$ components ranked by size are highlighted with distinct colors. The largest selected component is shown in red, while subsequent components receive different colors. These colors serve only as region identifiers rather than defect labels, and unselected failed dies remain visible in white.

We consider two forms of $Z$: Single-region highlights the largest component ($K=1$), whereas Multi-region highlights the five largest components ($K=5$). These alternatives control the scope of explicitly marked regional evidence without changing the underlying wafer map. The construction is fixed across different wafers.

Given $Z$, the textual prompt connects the highlighted visual evidence to the six class definitions. For each class $c\in\mathcal{Y}$, we define a structured semantic rule $r_c=(M_c,A_c,E_c)$, where $M_c$ describes the characteristic morphology, $A_c$ specifies its spatial anchor on the wafer, and $E_c$ distinguishes the class from neighboring interpretations. Section~\ref{sec:semantic} develops these three rule components. Let $\pi=(\pi_1,\ldots,\pi_{|\mathcal{Y}|})$ denote a permutation of the class labels that determines the presentation order of the rules in the textual prompt. This order encodes an intended priority for cases with overlapping evidence. The domain rationale for the selected order is developed in Section~\ref{sec:priority}.

The complete textual prompt is constructed as
\begin{equation}
P(Z;\mathcal{R},\pi)
=L(Z)\oplus\left[\bigoplus_{j=1}^{|\mathcal{Y}|}T\!\left(r_{\pi_j}\right)\right]\oplus C,
\qquad r_c=(M_c,A_c,E_c).
\label{eq:prompt-construction}
\end{equation}
where $L(Z)$ is the legend associated with the current visual representation, $T$ verbalizes a structured class rule in natural language, $C$ specifies the allowed labels and response format, and $\oplus$ denotes textual concatenation. Thus, the middle term preserves the semantic content of every class rule while arranging the rules according to $\pi$. The formulation makes explicit the two complementary design choices developed below: the internal construction of each rule $r_c$ and the presentation order assigned to the rule sequence. The prompt comprises a visual legend, the six rules summarized in Table~\ref{tab:rules} and arranged in the fixed order Donut $>$ Scratch $>$ Edge $>$ Random $>$ Center $>$ Local, and an instruction to return exactly one label. For Local and Scratch, it emphasizes the largest component; for Edge and Random, the global defect distribution. The same wording is used for all inputs within each visual condition. For every input, the same rule set $\mathcal{R}$ and selected presentation order $\pi$ are used without access to ground-truth labels or dataset-specific adaptation. Sections~\ref{sec:semantic} and~\ref{sec:priority} develop the rule semantics and presentation-order design represented in Eq.~\eqref{eq:prompt-construction}, respectively.

\subsection{Class-Semantic Rule Construction}
\label{sec:semantic}

Wafer defect classes are usually named by process engineers according to the spatial shape and location of defects, but the names alone are insufficient to constrain zero-shot LMM decisions. Center, Donut, Edge, Local, Random, and Scratch are not six fully independent low-level visual labels. They share part of their spatial evidence. For example, a near-center defect may be interpreted as Center because of its location, but as Donut because of its hollow structure; a local defect cluster close to the edge may be interpreted as Edge or Local. If the prompt only lists class names, the model may treat such shared evidence as final class evidence. We therefore rewrite each class as a semantic rule, so that the prompt states what should be observed, where it should be observed, and why neighboring classes should be rejected. Accordingly, this section defines the internal structure of each rule $r_c$ in Eq.~\eqref{eq:prompt-construction}, specifying what visual evidence supports class $c$ and what evidence separates it from its closest alternatives.

As introduced in Eq.~\eqref{eq:prompt-construction}, each class rule is represented as $r_c=(M_c,A_c,E_c)$, comprising morphology, spatial anchoring, and an exclusion boundary. Morphology $M_c$ characterizes the dominant geometric structure of a defect, such as a solid cluster, hollow ring, elongated trace, edge-following arc, or broadly distributed pattern. Spatial anchoring $A_c$ specifies the location of that structure relative to key wafer regions, including the center, interior, and circular boundary. The exclusion boundary $E_c$ distinguishes the target class from visually similar neighboring classes by identifying evidence that supports or rules out alternative interpretations. Table~\ref{tab:rules} summarizes these components.

\begin{table*}[!t]
  \caption{Class-semantic rules used in the textual prompt.}
  \label{tab:rules}
  \centering
  \footnotesize
  \setlength{\tabcolsep}{3pt}
  \renewcommand{\arraystretch}{0.98}
  \begin{tabularx}{\textwidth}{p{0.08\textwidth}p{0.26\textwidth}p{0.25\textwidth}X}
    \toprule
    Class & Positive morphology & Spatial anchor & Exclusion condition \\
    \midrule
    Center & Dense, compact, predominantly solid cluster & Centered or near wafer center & Exclude hollow, annular, rim-aligned, elongated, and wafer-wide patterns. \\
    Donut & Ring, hollow center, arc, crescent, C-shape, spiral, or partial annulus & Around or partly surrounding wafer center & Exclude solid central clusters without a clear opening. \\
    Edge & Continuous or fragmented arc, peripheral band, rim-following segment, or edge ring & Attached to and aligned with the circular boundary & Boundary contact alone is insufficient; Exclude inward compact clusters and non-circumferential elongated traces. \\
    Local & Thick, compact, solid clump or polygonal aggregation & Non-central interior, or boundary-near but extending inward & Exclude rim-following arcs, wafer-wide patterns, central clusters, and elongated traces. \\
    Random & Large block-like regions with broad interior coverage and little continuous gray area & Across much of the interior without rim confinement & Exclude local clusters, annular structures, elongated traces, and rim-aligned arcs or rings. \\
    Scratch & Long, narrow, line-like, stripe-like, diagonal, or aligned broken chain & Any wafer location & Exclude compact blobs with short tails and arcs whose principal direction follows the rim. \\
    \bottomrule
  \end{tabularx}
\end{table*}

\begin{figure}[t]
  \centering
  \includegraphics[width=0.88\columnwidth]{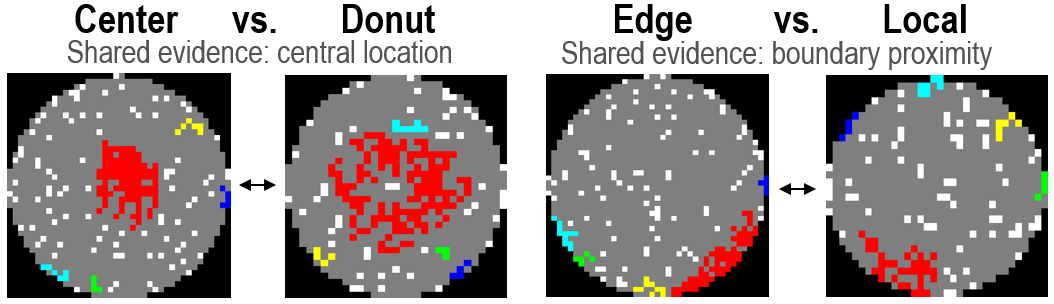}
  \caption{Visual grounding of two pairwise exclusion boundaries in the class-semantic rules. Center and Donut differ in solid versus hollow morphology; Edge and Local differ in rim alignment versus inward compactness.}
  \Description{Pairwise visual comparison of Center versus Donut and Edge versus Local, showing the shared location cues and the contrasting exclusion boundaries.}
  \label{fig:boundaries}
\end{figure}

The rules use qualitative morphology and spatial relations rather than dataset-specific thresholds or sample-level information. Figure~\ref{fig:boundaries} grounds two representative pairwise rules in the Multi-region visual representation used by the LMM. In the Center-Donut pair, both maps contain defect evidence near the wafer center. Their class assignments therefore cannot be determined from central location alone. The Center example contains a predominantly solid central mass, whereas the Donut example contains a meaningful hollow or annular structure in the dominant central pattern. The exclusion condition of the Center rule consequently prevents secondary surrounding regions from being interpreted as a Donut when the main central structure remains solid.

The Edge-Local pair illustrates a second shared-evidence boundary. Both examples contain defect regions near the wafer boundary, so boundary contact alone is not sufficient to support an Edge prediction. In the Edge example, the dominant defect structure follows the circular wafer rim as an arc-like segment. In the Local example, the dominant structure forms a compact clump that extends inward from the boundary rather than following its circumference. This distinction corresponds directly to the exclusion conditions in Table~\ref{tab:rules}: Edge rejects inward compact clusters, whereas Local rejects rim-following arcs and peripheral bands.

Together, the positive morphology, spatial anchor, and exclusion condition link each class decision to explicit visual evidence and distinguish it from the closest competing interpretations.

On the image side, connected-component marks identify the regions to which the rules refer. A single marked region presents a focused view of the largest component, whereas multiple marked regions present broader coverage when several components may be relevant. Their relative usefulness can vary across defect patterns according to the number, spatial arrangement, and diagnostic importance of the connected regions. We therefore compare different marking configurations to examine how the amount of highlighted visual evidence affects pattern-specific and overall performance.

\subsection{Rule-Ordered LMM Classification}
\label{sec:priority}

Section~\ref{sec:semantic} defines the semantic content of each rule $r_c$, but some wafer maps may remain partially compatible with multiple class descriptions. For example, a central defect may exhibit both Center-like location and Donut-like topology, while a thick boundary cluster may retain cues of both Local and Edge. We encode an intended priority for such cases through the permutation $\pi$ in Eq.~\eqref{eq:prompt-construction}. Specifically, $\pi$ determines the presentation sequence $T(r_{\pi_1}),\ldots,T(r_{\pi_{|\mathcal{Y}|}})$ in the textual prompt while leaving the semantic content of every $r_c$ unchanged. The prompt instructs the frozen LMM to consider earlier rules first when multiple descriptions appear compatible.

Because the LMM processes the complete image-text input jointly, this order provides an instruction-level decision bias rather than a deterministic symbolic decision tree.

The ordering of class-specific rules is treated as a controlled component of the prompt. This design is motivated by documented order sensitivity of LLMs: even when candidate content remains unchanged, reordering answer options or other input elements can alter model predictions, particularly when the model is uncertain among competing alternatives \cite{pezeshkpour2024}. Similar effects have been observed in closed-source and API-based LLMs across paraphrasing, relevance judgment, and multiple-choice question answering \cite{guan2025}. A fixed rule-presentation order therefore controls positional variation as an explicit input condition in frozen-LMM classification.

This control is especially important in the zero-shot setting, where no target-domain examples or parameter updates are available to help the model resolve ambiguity among visually similar classes. We use the fixed order Donut $>$ Scratch $>$ Edge $>$ Random $>$ Center $>$ Local as a domain-informed presentation sequence. The order is motivated by pairwise boundary relations among visually overlapping classes. Donut precedes Center because annular structure is more diagnostic than central location alone, while Scratch precedes Edge because non-circumferential elongation should be considered before boundary contact. Edge precedes Random so that rim-following arcs or rings are distinguished before broad interior coverage, and Random precedes Center and Local so that wafer-wide coverage is considered before compact aggregation. Finally, Center precedes Local because central anchoring is more specific than a general non-central compact region. These relations motivate the selected sequence but do not uniquely determine the relative positions of every class pair. The remaining placements are fixed as a consistent presentation convention and are retained across all inputs and datasets. The selected sequence is evaluated as a fixed prompt condition rather than as a globally optimal ordering.

The selected order leaves the definitions of Center, Donut, Edge, Local, Random, and Scratch unchanged and carries no assumption about their relative frequency or importance. Instead, its presentation sequence encodes an intended interpretive priority for cases in which a wafer map exhibits partially overlapping cues from multiple classes. Accordingly, the order is evaluated as a controlled prompt-presentation condition rather than as a deterministic decision tree. Algorithm~\ref{alg:classification} summarizes how the visual prompting interface and the ordered semantic rules are combined into an image-text input for the frozen LMM.

\begin{algorithm}[t]
\caption{Training-Free Wafer Map Classification}
\label{alg:classification}
\begin{algorithmic}[1]
\Require Wafer maps $\Phi$, frozen LMM $F_\theta$, and number of highlighted components $K$
\Ensure Predicted labels $\{\hat{y}_i\}_{\phi_i\in\Phi}$
\State Construct the six semantic rules and fix their presentation order.
\ForAll{wafer map $\phi_i\in\Phi$}
  \State Highlight the $K$ largest connected defect components.
  \State Combine the prompted map, visual legend, and ordered rules.
  \State $\hat{y}_i\gets F_\theta$(image-text pair).
\EndFor
\State \Return $\{\hat{y}_i\}_{\phi_i\in\Phi}$.
\end{algorithmic}
\end{algorithm}

\section{Experiments}

We evaluate controlled zero-shot prompt variants under a fixed frozen-LMM setting. Prompt templates were fixed before evaluation and reused across all samples and both datasets without dataset-specific modification. The exact same semantic rules and rule order were used for both datasets; legends and output constraints were reused for every map within each visual condition. Following the visual-prompt configuration in \cite{cao2026}, Raw preserves the original defect distribution without component overlays, Single-region marks one major connected defect region ($K=1$), whereas Multi-region marks five major connected defect regions ($K=5$). Neither the images nor the labels in Mixed-WM38 \cite{wang2020} and WM811K \cite{wu2015} are used for training, fine-tuning, or few-shot prompting before inference. When visual prompting is applied, it only marks connected defect regions according to Section~\ref{sec:overview}, with a legend that explains the visible marks.

All experiments use the frozen gpt-4o-2024-08-06 model through the OpenAI API \cite{hurst2024}, with the temperature fixed at 0.1. Each wafer map is submitted once as an image-text pair; thus, the reported results reflect single-query inference and do not characterize query-to-query variation. We report per-class accuracy, macro accuracy, defined as the unweighted mean of the six per-class accuracies, and overall accuracy, defined as the proportion of correctly classified maps in the complete evaluation set. Per-class accuracy is computed within each ground-truth class. Center, Donut, Edge, Local, Random, and Scratch will be abbreviated as C, D, E, L, R, and S, respectively.

\subsection{Mixed-WM38}

Mixed-WM38 \cite{wang2020} contains approximately 38,000 wafer maps covering normal wafers, eight single-pattern defect categories, and mixed-type defects. For the six-class evaluation considered in this study, we retain all single-pattern wafer maps whose source labels correspond to Center, Donut, Edge, Local, Random, or Scratch. The resulting evaluation set comprises 6,866 maps, including 1,000 maps for each of Center, Donut, Local, and Scratch, 2,000 maps for Edge by combining Edge\_Loc and Edge\_Ring, and 866 maps for Random. The original dataset labels are used exclusively to compute the per-class, macro, and overall accuracies reported in Tables~\ref{tab:mixed} and~\ref{tab:order}. They are not used to construct the prompts, adapt the model, or select predictions.

We first examine how the presentation and scope of regional evidence affect multi-class classification. Table~\ref{tab:mixed} compares four prompting conditions on Mixed-WM38: the unmodified wafer map, a single-region representation, a multi-region representation, and a names-only prompt. Regional visual marking produces a substantial improvement over the raw image representation. Single-region prompting increases the overall accuracy from 37.5\% to 93.3\% and the macro accuracy from 44.6\% to 92.4\%. Multi-region prompting further improves these values to 93.8\% and 92.9\%, respectively, and therefore achieves the best overall and macro accuracy in this comparison. Single-region performs better for Center and Local, whereas Multi-region provides higher accuracy for Donut, Edge, and Random by preserving additional spatial evidence from secondary components. When the semantic descriptions are removed and only class names are provided, the overall accuracy decreases to 63.6\% and the macro accuracy decreases to 58.2\%. This reduction indicates that class names alone do not sufficiently specify the morphology, spatial location, and exclusion boundaries required for fine-grained classification.

\begin{table}[t]
  \caption{Accuracy (\%) on Mixed-WM38.}
  \label{tab:mixed}
  \centering
  \footnotesize
  \setlength{\tabcolsep}{2.5pt}
  \begin{tabularx}{\columnwidth}{@{}X*{8}{r}@{}}
    \toprule
    Method & C & D & E & L & R & S & Macro & Overall \\
    \midrule
    Raw & 9.5 & 97.1 & 2.8 & 11.7 & \textbf{97.9} & 46.8 & 44.6 & 37.5 \\
    Single & \textbf{99.6} & 97.3 & 95.6 & \textbf{98.0} & 70.1 & \textbf{94.0} & 92.4 & 93.3 \\
    Multi & 97.6 & \textbf{98.7} & \textbf{96.8} & 92.1 & 78.5 & 93.9 & \textbf{92.9} & \textbf{93.8} \\
    Names-only & 69.7 & 39.4 & 90.9 & 64.2 & 24.7 & 60.5 & 58.2 & 63.6 \\
    \bottomrule
  \end{tabularx}
\end{table}

\subsection{Rule-presentation order}

We next isolate the effect of rule presentation by keeping the Multi-region images, visual legend, class-rule wording, and output format unchanged while varying only the rule-presentation order of the six rule blocks. The default sequence is D$>$S$>$E$>$R$>$C$>$L. The eight targeted variants modify this sequence to examine boundaries between visually related classes: Center-first uses C$>$D$>$S$>$E$>$R$>$L; Edge-before-Scratch uses D$>$E$>$S$>$R$>$C$>$L; Local-before-Edge uses D$>$S$>$L$>$E$>$R$>$C; Local-before-Scratch uses D$>$L$>$S$>$E$>$R$>$C; Random-before-Edge uses D$>$S$>$R$>$E$>$C$>$L; Random-last uses D$>$S$>$E$>$C$>$L$>$R; Edge-Local-Scratch uses D$>$E$>$L$>$S$>$R$>$C; and Center-Local swap uses D$>$S$>$E$>$R$>$L$>$C. Because the semantic content remains identical across these variants, the comparison directly characterizes the sensitivity of the frozen LMM to rule-presentation order.

\begin{table}[t]
  \caption{Accuracy (\%) of rule-presentation orders on Mixed-WM38.}
  \label{tab:order}
  \centering
  \footnotesize
  \setlength{\tabcolsep}{2.5pt}
  \begin{tabularx}{\columnwidth}{@{}X*{8}{r}@{}}
    \toprule
    Variant & C & D & E & L & R & S & Macro & Overall \\
    \midrule
    Default & \textbf{97.6} & \textbf{98.7} & \textbf{96.8} & 92.1 & 78.5 & 93.9 & \textbf{92.9} & \textbf{93.8} \\
    Center-first & 92.1 & 96.3 & 94.0 & 93.0 & 76.5 & 91.1 & 90.5 & 91.3 \\
    Edge-before-Scratch & 94.5 & 97.7 & 92.3 & 93.7 & 77.5 & 87.0 & 90.4 & 91.0 \\
    Local-before-Edge & 96.1 & 97.5 & 89.0 & 94.0 & 71.7 & 91.0 & 89.9 & 90.1 \\
    Local-before-Scratch & 95.5 & 98.1 & 92.0 & \textbf{96.0} & 80.1 & 80.5 & 90.4 & 90.8 \\
    Random-before-Edge & 94.0 & 96.9 & 91.5 & 95.5 & 80.3 & 92.3 & 91.8 & 91.9 \\
    Random-last & 97.5 & 97.7 & 85.5 & 95.0 & \textbf{86.5} & \textbf{94.5} & 92.8 & 91.9 \\
    Edge-Local-Scratch & 97.0 & 98.0 & 91.8 & 94.5 & 84.5 & 77.5 & 90.5 & 90.8 \\
    Center-Local swap & 94.3 & 98.1 & 86.8 & \textbf{96.0} & 80.5 & 94.0 & 91.6 & 91.1 \\
    \bottomrule
  \end{tabularx}
\end{table}

The default order achieves the highest overall accuracy, reaching 93.8\%, with a macro accuracy of 92.9\%. All eight alternative orders reduce the overall accuracy to between 90.1\% and 91.9\%, although the effects are not uniform across classes. Placing Local before Edge increases Local accuracy from 92.1\% to 94.0\%, but decreases Edge accuracy from 96.8\% to 89.0\%. Placing Random last increases Random accuracy from 78.5\% to 86.5\%, while reducing Edge accuracy from 96.8\% to 85.5\%. Similarly, placing Local before Scratch improves Local accuracy to 96.0\%, but reduces Scratch accuracy from 93.9\% to 80.5\%. These non-uniform changes indicate that rule order acts as an instruction-level bias for resolving partially compatible class descriptions, rather than as a deterministic priority mechanism \cite{pezeshkpour2024}.

\subsection{WM811K}

WM811K \cite{wu2015} contains 172,950 pre-labeled wafer maps and is a large dataset collected from a real manufacturing environment. For testing, we randomly selected 9,000 maps for the six target patterns: 1,747 Center, 502 Donut, 3,430 Edge, 1,830 Local, 555 Random, and 936 Scratch maps. Edge again combines the two edge-pattern classes Edge\_Loc and Edge\_Ring.

\begin{table}[t]
  \caption{Accuracy (\%) on WM811K.}
  \label{tab:wm811k}
  \centering
  \footnotesize
  \setlength{\tabcolsep}{2.5pt}
  \begin{tabularx}{\columnwidth}{@{}X*{8}{r}@{}}
    \toprule
    Method & C & D & E & L & R & S & Macro & Overall \\
    \midrule
    Raw & 52.4 & 72.1 & 26.5 & 31.0 & \textbf{99.6} & 70.8 & 58.7 & 43.7 \\
    Single & \textbf{94.7} & 61.6 & 79.1 & \textbf{78.4} & 69.8 & 76.4 & 76.7 & 80.1 \\
    Multi & 91.3 & \textbf{73.1} & \textbf{81.0} & 76.4 & 74.8 & \textbf{80.9} & \textbf{79.1} & \textbf{81.3} \\
    \bottomrule
  \end{tabularx}
\end{table}

Table~\ref{tab:wm811k} shows that regional prompting remains effective on WM811K. Single-region and Multi-region improve overall accuracy from 43.7\% for Raw to 80.1\% and 81.3\%, respectively, while increasing macro accuracy from 58.7\% to 76.7\% and 79.1\%. The improvement in macro accuracy indicates that the overall gain is not attributable only to the relatively large Edge class. Multi-region achieves the best overall and macro accuracies, although Single-region remains more effective for Center and Local. This result is consistent with the evidence-scope trade-off observed on Mixed-WM38: a single highlighted component emphasizes a dominant compact region, whereas multiple highlighted components better preserve distributed or secondary evidence.

Nevertheless, Multi-region accuracy is lower on WM811K than on Mixed-WM38 by 12.5 percentage points overall and 13.8 percentage points in macro accuracy. The largest class-wise decrease occurs for Donut, followed by Edge, Local, and Scratch. As discussed in \cite{cao2026}, Mixed-WM38 contains uniformly sized wafer maps with visually dominant abnormal patterns, whereas WM811K contains wafer maps of varying sizes and many ambiguous or fragmented defect patterns. These can weaken the morphology required to distinguish neighboring classes, such as Donut from Center or Local, and Edge from Scratch or Local.

\begin{figure}[t]
  \centering
  \includegraphics[width=0.80\columnwidth]{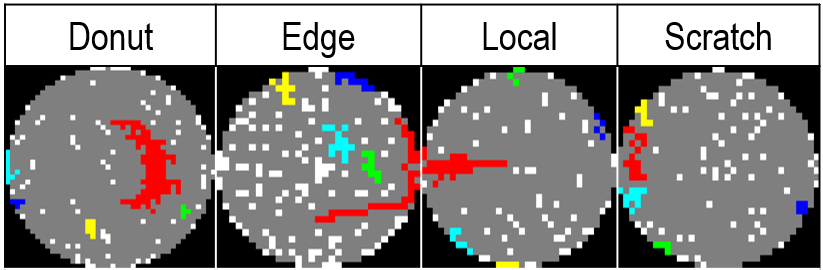}
  \caption{Representative WM811K samples with weak or ambiguous class-specific morphologies; column headings denote source labels.}
  \Description{Four representative WM811K wafer maps with weak or ambiguous class-specific morphologies. The source labels are Donut, Edge, Local, and Scratch. Colored overlays identify connected components rather than defect classes.}
  \label{fig:wm811k-ambiguity}
\end{figure}

Figure~\ref{fig:wm811k-ambiguity} illustrates these sources of error. Some source labels are difficult to reconcile with the dominant visual morphology: the Donut sample contains only an incomplete curved structure without a clear hollow center or annular topology, making the source label difficult to distinguish from Local. The Edge sample combines boundary contact with elongated structures and therefore provides evidence compatible with Edge and Scratch. The dominant region in the Local sample is long and narrow rather than compact, which makes it visually closer to Scratch. Conversely, the highlighted region in the Scratch sample is short and attached to the wafer boundary, with limited long-axis continuity, making Edge or Local plausible alternatives. These examples show that some WM811K wafer maps only weakly express the morphology associated with their source labels.

The results therefore suggest that the remaining performance gap arises from weaker morphological salience, greater within-class variation, and overlapping class boundaries in WM811K.

API inference required approximately 0.5--1 s per image in our environment. Although it is slower and more costly than local inference with a trained classifier, it removes the need for target-domain training before deployment.

\section{Conclusion}

This paper presents a training-free framework for zero-shot multi-class wafer map classification with a frozen large multimodal model. The framework combines connected-component highlighting with boundary-aware semantic rules to make defect regions, class characteristics, and exclusion boundaries explicit in the image-text input. A fixed rule-presentation order further provides a consistent instruction-level mechanism for resolving partially overlapping class descriptions. Experiments on two industrial datasets show that regional highlighting and complete semantic rules substantially improve classification over raw inputs and class names alone, while controlled order variants confirm that the placement of competing rules affects the resolution of ambiguous evidence. These findings show that prompt construction can serve as an explicit semantic decision specification for fine-grained wafer map recognition. The framework requires no labeled training examples, fine-tuning, or parameter updates, and its visual and textual specifications can be inspected and revised for new classification requirements. Future work will evaluate robustness across multimodal models and mixed-pattern wafer maps.

\end{document}